\documentclass[sigconf]{acmart}
\renewcommand\footnotetextcopyrightpermission[1]{} 
\AtBeginDocument{%
  \providecommand\BibTeX{{%
    \normalfont B\kern-0.5em{\scshape i\kern-0.25em b}\kern-0.8em\TeX}}}

\setcopyright{acmcopyright}
\copyrightyear{2018}
\acmYear{2018}
\acmDOI{10.1145/1122445.1122456}

\acmConference[Preprint]{Preprint}
\acmBooktitle{2023}

\usepackage{algorithm} 
\usepackage{algpseudocode}
\usepackage{multirow}
\makeatletter
\newcommand{\StatexIndent}[1][3]{%
  \setlength\@tempdima{\algorithmicindent}%
  \Statex\hskip\dimexpr#1\@tempdima\relax}
\usepackage{url} 
\usepackage{fancyvrb}
\usepackage{fancyhdr}
\usepackage{colortbl}
\usepackage{multicol}
  
\begin{document}

\title{Down-Sampled Epsilon-Lexicase Selection for Real-World Symbolic Regression Problems}

\author{Alina Geiger}
\affiliation{%
  \institution{Johannes Gutenberg University}
  \city{Mainz}
  \country{Germany}}
\email{ageige01@students.uni-mainz.de}

\author{Dominik Sobania}
\affiliation{%
  \institution{Johannes Gutenberg University}
  \city{Mainz}
  \country{Germany}}
\email{dsobania@uni-mainz.de}

\author{Franz Rothlauf}
\affiliation{%
  \institution{Johannes Gutenberg University}
  \city{Mainz}
  \country{Germany}}
\email{rothlauf@uni-mainz.de}

\renewcommand{\shortauthors}{Geiger et al.}

\begin{abstract}
Epsilon-lexicase selection is a parent selection method in genetic programming that has been successfully applied to symbolic regression problems. Recently, the combination of random subsampling with lexicase selection significantly improved performance in other genetic programming domains such as program synthesis. However, the influence of subsampling on the solution quality of real-world symbolic regression problems has not yet been studied. In this paper, we propose down-sampled epsilon-lexicase selection which combines epsilon-lexicase selection with random subsampling to improve the performance in the domain of symbolic regression. Therefore, we compare down-sampled epsilon-lexicase with traditional selection methods on common real-world symbolic regression problems and analyze its influence on the properties of the population over a genetic programming run. We find that the diversity is reduced by using down-sampled epsilon-lexicase selection compared to standard epsilon-lexicase selection. This comes along with high hyperselection rates we observe for down-sampled epsilon-lexicase selection. Further, we find that down-sampled epsilon-lexicase selection outperforms the traditional selection methods on all studied problems. Overall, with down-sampled epsilon-lexicase selection we observe an improvement of the solution quality of up to $85\%$ in comparison to standard epsilon-lexicase selection.
\end{abstract}



\maketitle

\vfill\eject

\section{Introduction}

Symbolic regression is a problem that involves finding a mathematical expression that describes a given data set \cite{Poli.2008}. The application possibilities of symbolic regression are wide and range from finance \cite{Chen.2012} to even medicine \cite{Virgolin.2018}. The most common approach to solve symbolic regression problems is genetic programming (GP) \cite{Koza.1992}, where candidate solutions (individuals) are evolved in an evolutionary process. To gradually improve the population during evolution, individuals are selected as parents based on their fitness (e.g., performance on the training data) for the next generation.

Traditional parent selection methods, such as tournament selection, use an aggregated fitness value to evaluate the quality of the individuals, causing a loss of information for the search \cite{Krawiec.2014}. In contrast, lexicase selection \cite{Spector.2012} selects individuals by considering the error on each training case separately. Prior work has shown that lexicase selection improves problem-solving performance in many GP application domains \cite{Helmuth.2014, Helmuth.2015, Moore.2017, Sobania.2023}. 

However, standard lexicase selection is not able to achieve the same performance improvements when applied to continuous-valued symbolic regression problems due to its strict pass condition as only candidates that perform best on each considered training case are selected. Therefore, $\epsilon$-lexicase selection has been proposed that relaxes the pass condition by introducing an $\epsilon$ threshold \cite{LaCava.2016, LaCava.2019}. $\epsilon$-lexicase selection has shown to perform well over a wide range of regression problems compared to traditional methods \cite{LaCava.2021}.

In other application domains of GP, such as program synthesis, the combination of random subsampling with lexicase selection significantly improved performance \cite{Hernandez.2019, Ferguson.2020}. In each generation, a random subset of training cases instead of all available training cases is used for all selection events, allowing more individuals to be evaluated with the same evaluation budget. However, despite its benefits, down-sampled lexicase selection is not appropriate for symbolic regression problems as it is based on standard lexicase selection. 

Therefore, we propose in this paper down-sampled $\epsilon$-lexicase selection and compare its performance with traditional selection methods on common real-world symbolic regression benchmarks. Furthermore, we analyze how this novel selection method influences the properties of the population over a GP run. 

Down-sampled $\epsilon$-lexicase selection uses a random subset of training cases in each generation for the selections with $\epsilon$-lexicase. The subsampling level $s$ defines the size of the subset and the $\epsilon$ value defines the strength of the filtering during the selection process. We analyze the influence of both parameters on behavioral diversity, the hyperselection rate, as well as on the ability to select specialists. The experiments are performed on six common symbolic regression benchmarks from the UCI machine learning repository \cite{Dua.2019}.
 
We find that down-sampled $\epsilon$-lexicase selection outperforms traditional selection methods on all studied symbolic regression problems. Furthermore, we find that behavioral diversity is reduced by using down-sampled $\epsilon$-lexicase selection compared to $\epsilon$-lexicase selection. 

Following this introduction, Sect.~\ref{sec:related_work} gives a brief overview of related work on the traditional selection methods analyzed in this study. Section~\ref{sec:method} describes our novel selection method, the used metrics, and the used benchmark problems. In Sect.~\ref{sec:experiments}, we describe our experimental setting and discuss the results before concluding the paper in Sect.~\ref{sec:conclusion}.

\section{Related Work}\label{sec:related_work}

Selection methods have a major impact on the evolutionary search process. Traditionally, selection methods are based on a compressed fitness value where the information of the error observed on all training cases is combined. A common fitness-based selection method is tournament selection, where in each selection event $k$ random individuals participate in a tournament and the best one is selected as a parent \cite{Poli.2008}. However, selecting individuals based on a single fitness-value leads to an information loss as the structural information of the training data is not considered \cite{Krawiec.2014}. 

A method that addresses this issue is lexicase selection \cite{Spector.2012}, where the performance on the individual training cases is considered instead of an aggregated fitness value. In detail, lexicase selection considers all individuals in the population as candidates for a selection event. After that, it iterates through the training cases, keeping only the candidates with the lowest error on the current case. This is repeated until either only one candidate remains or there are no more training cases. If more than one candidate remains, one of them is randomly chosen.

In recent years, especially in program synthesis, great progress has been made with lexicase selection compared to the common fitness-value based selection methods \cite{Forstenlechner.2017, Helmuth.2015, Helmuth.07082020, Saini.2020, Sobania.06262021}. Due to the success of lexicase selection, studies have been proposed that theoretically analyze the properties of lexicase selection \cite{Helmuth.2016c, Helmuth.2020b, Jansen.2018}.  

However, lexicase selection is not able to achieve the same performance improvements on regression problems, as its strict pass condition is not suitable for continuous-valued problems. Therefore, La Cava et al.~\cite{LaCava.2016} introduced $\epsilon$-lexicase selection that relaxes the pass condition of lexicase selection by introducing an $\epsilon$ threshold that also allows candidates to pass with an error within the $\epsilon$ threshold on each case. $\epsilon$-lexicase selection has been found to perform well in comparison to other regression approaches \cite{LaCava.2021}.

In the domain of program synthesis, the combination of subsampling and lexicase selection has been shown to improve performance over a wide range of problems \cite{Hernandez.2019, Ferguson.2020}. Subsampling means that each individual is only evaluated against a fraction of the training set, which overall enables the evaluation of more individuals with the same evaluation budget. In recent years, different variants of subsampling-based selection methods have emerged \cite{Hernandez.2019, Helmuth.2020c, Helmuth.2021, Ferguson.2020, Boldi.2023, Schweim.2022}.  

However, to our knowledge, no work so far studied the combination of $\epsilon$-lexicase selection and subsampling on common real-world regression problems and analyzed its influence on population dynamics.

\section{Methodology}\label{sec:method}
 
In order to study whether the performance benefits that random subsampling has shown for other GP application domains can be transferred to the domain of symbolic regression, we introduce down-sampled $\epsilon$-lexicase selection. In this section, we describe this novel selection method and explain the metrics used to analyze its influence on the evolutionary search. In addition, we present the benchmark problems used for evaluation.

\subsection{Down-Sampled $\epsilon$-Lexicase Selection}\label{sec:down_ep_lex}

Down-sampled $\epsilon$-lexicase selection is a combination of random subsampling \cite{Goncalves.2012} and $\epsilon$-lexicase selection \cite{LaCava.2016, LaCava.2019}, which means that in each generation a random subset $S$ of the training cases $t \in T$ is used for all selections with $\epsilon$-lexicase. This reduces the number of evaluations per generation as each individual is only evaluated against a fraction of the training set $T$. By allocating the saved evaluations to a longer search or to larger population sizes, more individuals can be evaluated with the same evaluation budget \cite{Hernandez.2019, Helmuth.2021}. 

The down-sampled $\epsilon$-lexicase algorithm is shown in Algorithm~\ref{algo:down-ep-lex-math} for all $|P|$ selection events in a generation for a population $P$ and a training set $T$.

\begin{algorithm}[H]
	\caption{Down-Sampled $\epsilon$-Lexicase Selection}
	\label{algo:down-ep-lex-math}
	\begin{algorithmic}[1]
\State $S$ := random subset of $T$ of size $s$\label{line:init_subset}
\State $P'$ := $\emptyset$ \label{line:init_offspring}
\For{$j$ in $|P|$}\label{line:loop_gen}
    \State $C$ := all individuals $i \in P$\label{line:init_candidates}
    \State $S'$ := randomly shuffled $S$\label{line:init_cases}
        \While {$|S'|$ $>$ 0 and $|C|$ $>$ 1}\label{line:loop_sel}
            \State $t$ := first case of $S'$
            \State $\lambda_t$ := $e^*_t + \epsilon$ \label{line:threshold}
            \State $C$ := all candidates with $e_t <= \lambda_t$\label{line:pass_condition}
            \State \textbf{remove} $t$ from $S'$
        \EndWhile\label{line:end_loop_sel}
    \State $p$ := random choice from $C$\label{line:chose_offspring}
    \State  $P'$ := add $p$ to $P'$\label{line:update_offspring}
\EndFor\label{line:end_loop_gen}
\State  \textbf{return} $P'$\label{line:return_offspring}
	\end{algorithmic} 
\end{algorithm}

In each generation, a random subset of the training set $T$ is chosen (line \ref{line:init_subset}). The subsampling level $s$ defines the size of the subset. For example, for $s = 0.1$ the random sample consists of $10\%$ of the training set $T$. Based on this subset, all $|P|$ selection events in a generation are performed (line \ref{line:loop_gen}-\ref{line:end_loop_gen}). For each selection event, all individuals $i \in P$ form the candidate pool $C$ and all training cases $t \in S$ are randomly ordered to form the case pool $S'$ (line \ref{line:init_candidates}-\ref{line:init_cases}). The algorithm then iterates through the training cases $t \in S'$, always keeping only the candidates with an error $e_t$ that is less than or equal to the threshold $\lambda_t$ on case $t$. The threshold $\lambda_t$ is defined as the lowest error $e^*_t$ among all candidates in the current candidate pool $C$ plus $\epsilon$. This is repeated until only one candidate remains or there are no more training cases (line \ref{line:loop_sel}-\ref{line:end_loop_sel}). If there is more than one candidate left, one is randomly chosen from the remaining candidate pool $C$ as a parent (line \ref{line:chose_offspring}). At the end, all $|P|$ individuals  selected are returned (line \ref{line:return_offspring}). 

The parameter $\epsilon$ can be either defined as an absolute static parameter or it can be dynamically determined in each iteration step with respect to the performance of the current candidate pool $C$ on case $t$. The variant with a static $\epsilon$ allows us to analyze the influence of different values of $\epsilon$. However, the optimal $\epsilon$ value is problem dependent \cite{LaCava.2016}. Therefore, the advantage of the dynamic approach is that it does not require the user to tune $\epsilon$.

For the dynamic calculation of $\epsilon$, we refer to the approach suggested by La Cava et al.~\cite{LaCava.2019}.
The performance of the candidate pool $C$ on case $t$ is estimated using the median absolute deviation \cite{PhamGia.2001}. Formally, $\epsilon_t$ for case $t$ is defined as
\begin{equation}\label{MAD}
\epsilon_t = \mathrm{median}(|\textbf{e}_t - \mathrm{median}(\textbf{e}_t)|),
\end{equation}
where $\textbf{e}_t$ is a vector of all error values on case $t$ across the candidates in the current candidate pool $C$ \cite{LaCava.2019}.

\subsection{Metrics}\label{sec:metrics}

Besides the performance, we analyze the influence of down-sampled $\epsilon$-lexicase selection on relevant properties of the population. 
Prior work has found that lexicase selection and $\epsilon$-lexicase selection produce populations with greater diversity than tournament selection, which is assumed to be one of the reasons for the successful search with lexicase-based selection methods \cite{Helmuth.2014, Helmuth.2016, Helmuth.2016c, LaCava.2016}. 

However, Helmuth et al. \cite{Helmuth.2016b} have found that lexicase selection produces hyperselection events which lead to drops in diversity. Hyperselection events are characterized by a single individual being selected in a large fraction of the selection events in a generation \cite{Helmuth.2016b}. It has been found that individuals with dominant error vectors in the population are selected more frequently with lexicase selection than with tournament selection \cite{Helmuth.2016b}. Consequently, we analyze the number of hyperselections to further explain the observed diversity using down-sampled $\epsilon$-lexicase selection compared to traditional selection methods. 

Apart from population diversity, the selection of specialists has been found to be one of the main reasons for the performance improvement of lexicase selection compared to tournament selection \cite{Helmuth.2020b, Helmuth.2019}.
By considering the performance on individual training cases rather than an aggregated fitness value, lexicase selection often selects individuals that perform extremely well on certain problem parts while performing poorly on average \cite{Helmuth.2020b}. This can be beneficial for the evolution as specialist are able to solve parts of the problem that are difficult for others.   

We measure diversity, hyperselection and specialist selection by means of the following metrics:  

\begin{itemize}
    \item \textbf{Diversity}: We focus on differences regarding the output of individuals to measure diversity. Therefore, we use the behavioral diversity metric which has been found to correlate with problem-solving performance \cite{Jackson.2010}. Behavioral diversity measures the percentage of distinct behavioral vectors in a population, where each behavioral vector captures the output $\hat{y}_t $ of an individual  $i \in P$ on each training case $t \in T$. 
    
    \item \textbf{Hyperselection}: 
    The number of hyperselected individuals per generation is measured by counting how many times the same individual is selected \cite{Helmuth.2016b}. We consider two individuals the same if the string representations of their trees are the same. 
    We report the number of hyperselections at the 1\%, 5\%, and 10\% level as proposed by Helmuth et al. \cite{Helmuth.2016b}. The level indicates the percentage of selection events in which an individual is at least selected in a generation. For example, an individual that is selected in 50 out of 500 selection events in a generation is defined as hyperselected at the 10\% level (as well as at the 1\% and 5\% levels) \cite{Helmuth.2016b}. 
    
    \item \textbf{Specialist selection}: 
    To analyze the ability of down-sampled $\epsilon$-lexicase selection to select specialists, we follow the experimental design suggested by Pantridge et al. \cite{Pantridge.2018}. We compare the mean number of used training cases per selection event and the mean rank of the selected individuals in each generation. The rank of an individual is defined by its average performance compared to the other individuals in the population. E.g., the individual with the best average fitness has rank 1, the second best has rank 2 and so on. If two individuals have the same average fitness they share a rank.
    An individual with a poor rank that is selected using only a few training cases is considered to be a specialist, because this indicates that it outperformed others on a small subset of training cases while performing poorly on average.
\end{itemize}

\subsection{Benchmark Problems}\label{sec:rproblems}

We evaluate down-sampled $\epsilon$-lexicase selection and the traditional selection methods on six real-world regression problems taken from the UCI machine learning repository \cite{Dua.2019}. The selected problems are commonly used in the GP literature \cite{LaCava.2019, Virgolin.2021, Martins.2018} and cover a diverse set of applications ranging from the estimation of house prices to the hydrodynamic performance of sailing yachts. Table~\ref{tab:problems} presents the dimension and the number of observations for all considered datasets. 

\definecolor{lightgray}{gray}{0.9}
\begin{table}[H]
\renewcommand{\arraystretch}{1.18}
  \begin{center}
    \caption{Dimension and number of observations for all considered datasets.}
    \label{tab:problems}
    \begin{tabular}{lrr}
     \toprule 
      \textbf{Problem} & \textbf{Dimension} & \textbf{Observations}\\
      \midrule
      \rowcolor{lightgray}
      Airfoil \cite{Brooks.1989} & 5 & 1,503  \\
      Concrete \cite{Yeh.1998} & 8 & 1,030  \\
      \rowcolor{lightgray}
      ENC \cite{Tsanas.2012} &  8 & 768 \\
      ENH \cite{Tsanas.2012} &  8 & 768 \\
      \rowcolor{lightgray}
      Housing \cite{Harrison.1978} &  13 & 506 \\
      Yacht \cite{Gerritsma.1981} & 6 & 308  \\
      \bottomrule 
    \end{tabular}
  \end{center}
\end{table}

The number of observations range from 308 for the Yacht dataset to Airfoil with 1503 observations and the dimensions range from 5 features for the Airfoil dataset to Housing with 13 features.

\section{Experiments}\label{sec:experiments}

We present the experimental setting and discuss the achieved results.

\subsection{Experimental Setting}\label{sec:exp_setting}

In our experiments, we compare the static and the dynamic variant of down-sampled $\epsilon$-lexicase selection to tournament selection, lexicase selection and $\epsilon$-lexicase selection (dynamic variant as described in \cite{LaCava.2019}).
We use the evolutionary computation framework DEAP \cite{Fortin.2012} for our experiments. Table \ref{tab:gpsetting} summarizes the parameter settings of the GP approach. 

\begin{table}[H]
  \begin{center}
  \renewcommand{\arraystretch}{1.18}
    \caption{GP parameter settings}
    \label{tab:gpsetting}
    \begin{tabular}{lr}
     \toprule 
      \textbf{Parameter} & \textbf{Value} \\
      \midrule
      \rowcolor{lightgray}
      Population size & 500 \\
      Primitive set & \{\textbf{x}, ERC, +, -, *, AQ, sin, cos\} \\
      \rowcolor{lightgray}
      ERC range & [-1,1] \\
      Initialization method & Ramped half-and-half \\
      \rowcolor{lightgray}
      Maximum tree depth & 17 \\
      Crossover probability & 0.8 \\
      \rowcolor{lightgray}
      Mutation probability & 0.05 \\
      Max. generations for runs & \multirow{2}{*}{50} \\
      using all training cases \\
      \rowcolor{lightgray}
      Runs & 30 \\
      \bottomrule 
    \end{tabular}
  \end{center}
\end{table}

As functions, we use addition, subtraction, multiplication, sine, cosine and the analytic quotient (AQ) \cite{Ni.2013} defined by 
\begin{equation}\label{aq}
\mathrm{AQ}(a,b) = \frac {a}{\sqrt{1+b^2}} ,
\end{equation}
where $a$ is the dividend and $b$ is the divisor. The terminals are the features $\textbf{x}$ of the considered problem and an ephemeral random constant (ERC) ranging from $-1$ to $1$.

We initialize the population with the ramped half-and-half method with tree depths ranging from 0 to 4. The maximum tree depth is limited to 17 \cite{Koza.1992}. We apply crossover with a probability of $0.8$ and mutation with a probability of $0.05$.  

For each problem, we divided the observations into training, validation, and test cases (70/15/15). The training cases are used to evaluate the quality of the individuals in the selection process (during the training phase). For tournament selection, we use the mean squared error (MSE) 
\begin{equation}\label{MSE}
\mathrm{MSE}(T) = \frac {1}{|T|} \sum_{t \in T} (y_t - \hat{y}_t)^2
\end{equation}
as the fitness function which averages the squared error between the real output $y_t$ and the output of an individual $\hat{y}_t$ for all $|T|$ training cases $t \in T$. However, lexicase-based selection methods evaluate individuals on each training case separately. Therefore, we define for those methods the error in terms of the squared error on each training case. 
To avoid choosing an individual as the solution of a GP run that is overfitted to the training cases, we evaluate the performance of all individuals that ever lived in the population on the validation cases. The best one is chosen as the solution and is evaluated on the unseen test cases in terms of its MSE. 

We compare the performance of the selection methods based on a given evaluation budget. 
The evaluation budget for each problem is defined as the number of training cases multiplied by the population size and the number of generations. We set the population size to 500 and we limit the number of generations for runs using all training cases to 50. Down-sampled $\epsilon$-lexicase selection uses only a subset of training cases each generation, therefore we allocate the saved evaluations to a longer search. This means that for a subsampling level $s = 0.1$ we run for 500 generations, for $s=0.2$ for 250 generations, and for $s=0.3$ for 167 generations (rounded).

For the parameterized selection methods, we have to specify values for each parameter. For tournament selection we use a tournament size $k = 5$. For down-sampled $\epsilon$-lexicase selection, we analyze $\epsilon$ values of $0$, $0.5$, $1$, and $5$ \cite{LaCava.2016} and subsampling levels $s$ of $0.1$, $0.2$, and $0.3$. For each combination of problem, selection method, and parameter setting we perform $30$ runs.

For all lexicase-based selection methods, we make an adjustment that does not change the selection probability of the individuals but lowers the case usage. In particular, we implement a termination condition that breaks the loop in the selection process as soon as all remaining candidates have identical error vectors. This prevents running through the remaining cases as this would not further reduce the candidate pool \cite{Helmuth.2020b}.

In each generation, we track the population metrics as described in Sect.~\ref{sec:metrics}. However, the runs using down-sampled $\epsilon$-lexicase selection train for more generations than the other selection methods using the same evaluation budget. Therefore, we show the results over the number of evaluations instead of the number of generations to ensure a fair comparison.

\subsection{The Influence of the Parameters of Down-Sampled $\epsilon$-Lexicase}\label{sec:param_setting}

To gain a deeper understanding of down-sampled $\epsilon$-lexicase selection, we analyze the influence of different $\epsilon$ values and subsampling levels on the performance, as well as on population diversity, hyperselection rates and specialist selection. The metrics used to measure the properties of the population are described in Sect.~\ref{sec:metrics}. The results are similar across all problems, therefore, due to space limitations, we exemplarily show the results for the ENC problem.   

\begin{figure}[h]
    \centering
    \includegraphics[scale = 0.555]{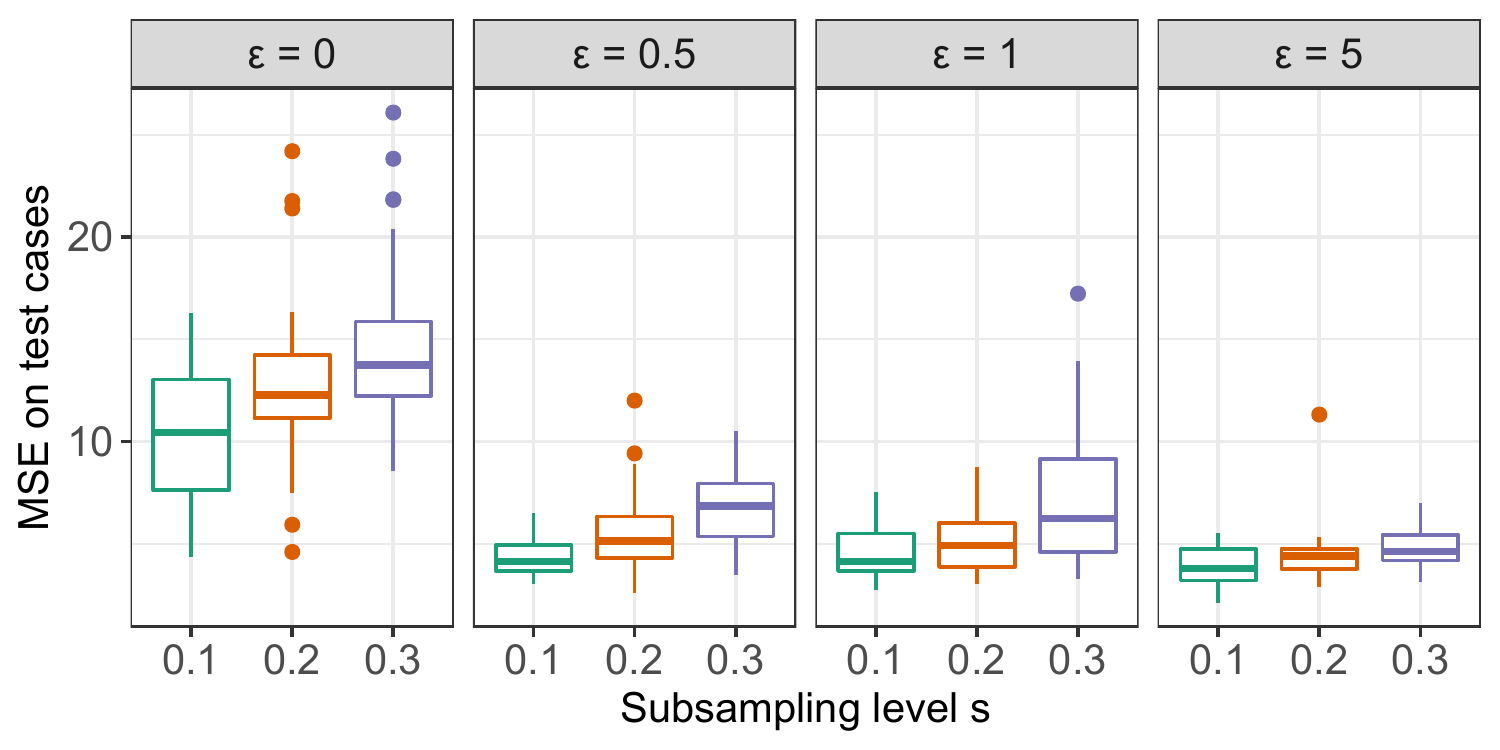}
    \caption{MSE on the test cases for all parameter settings of down-sampled $\epsilon$-lexicase selection for the ENC problem.}
    \label{fig:sub_param_tuning_enc}
\end{figure}

Figure \ref{fig:sub_param_tuning_enc} shows the distribution of the MSE on the unseen test cases for different subsampling levels $s$ and different $\epsilon$ values. We can observe that a lower subsampling level performs better. For example, for $\epsilon = 1$, a subsampling level $s = 0.3$ achieves only a median MSE of 6.227 on the test cases, while the median MSE with a subsampling $s = 0.1$ is 4.126. We assume that the improved performance with lower subsampling levels is due to the possibility to evaluate more individuals with the same evaluation budget as this has been found to be the main advantage of random subsampling in the domain of program synthesis \cite{Helmuth.2021}. In addition, we can observe that $\epsilon > 0$ performs better, which supports the motivation to use $\epsilon$-lexicase selection instead of lexicase selection for real-world regression problems.

\begin{figure}[h]
    \centering
    \includegraphics[scale=0.555]{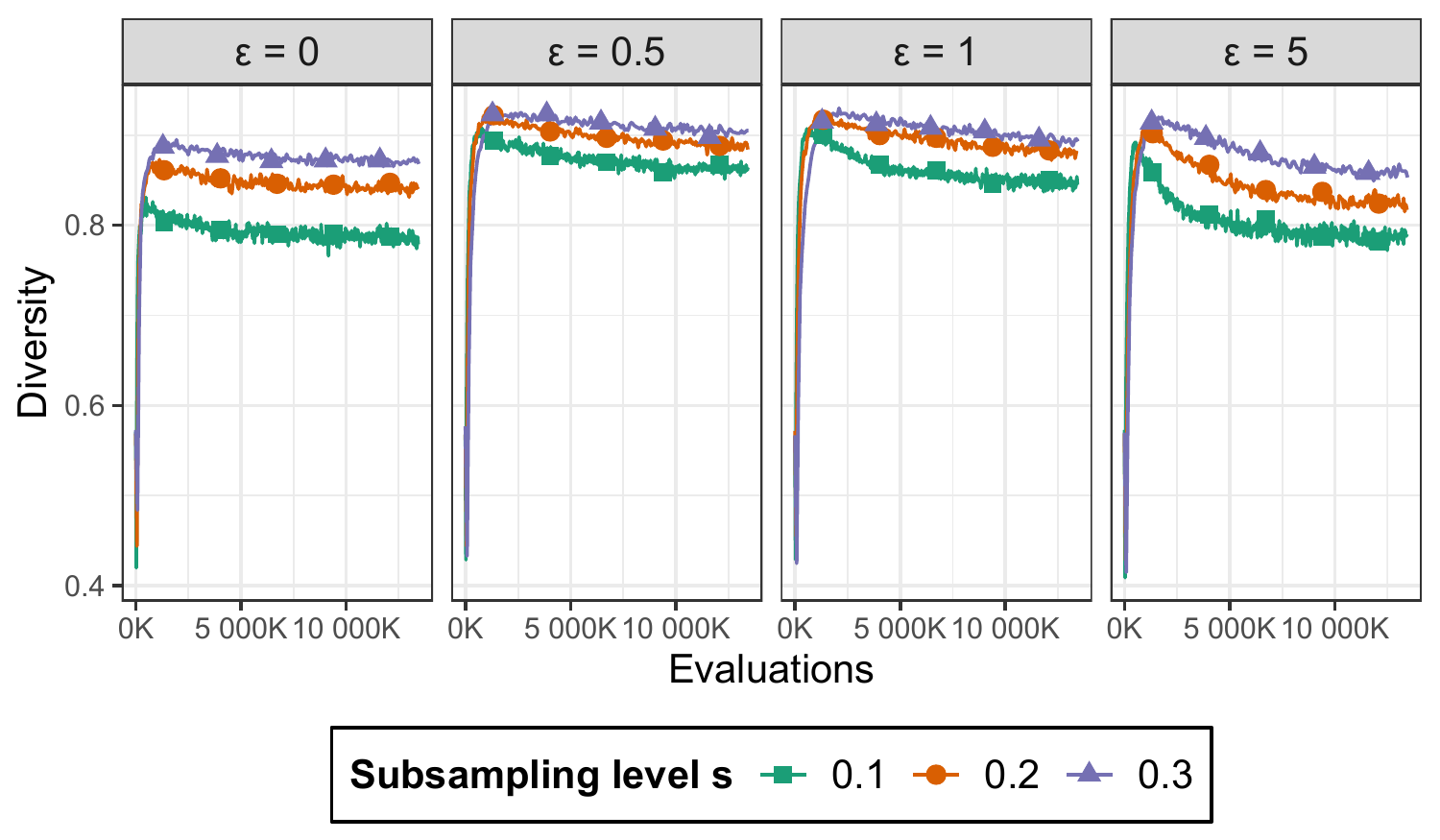}
    \caption{Behavioral diversity over the number of evaluations for all parameter settings of down-sampled $\epsilon$-lexicase selection for the ENC problem. The results are averaged over 30 runs.}
    \label{fig:sub_param_div_enc}
\end{figure}

\begin{figure}[h]
    \centering
    \includegraphics[scale=0.555]{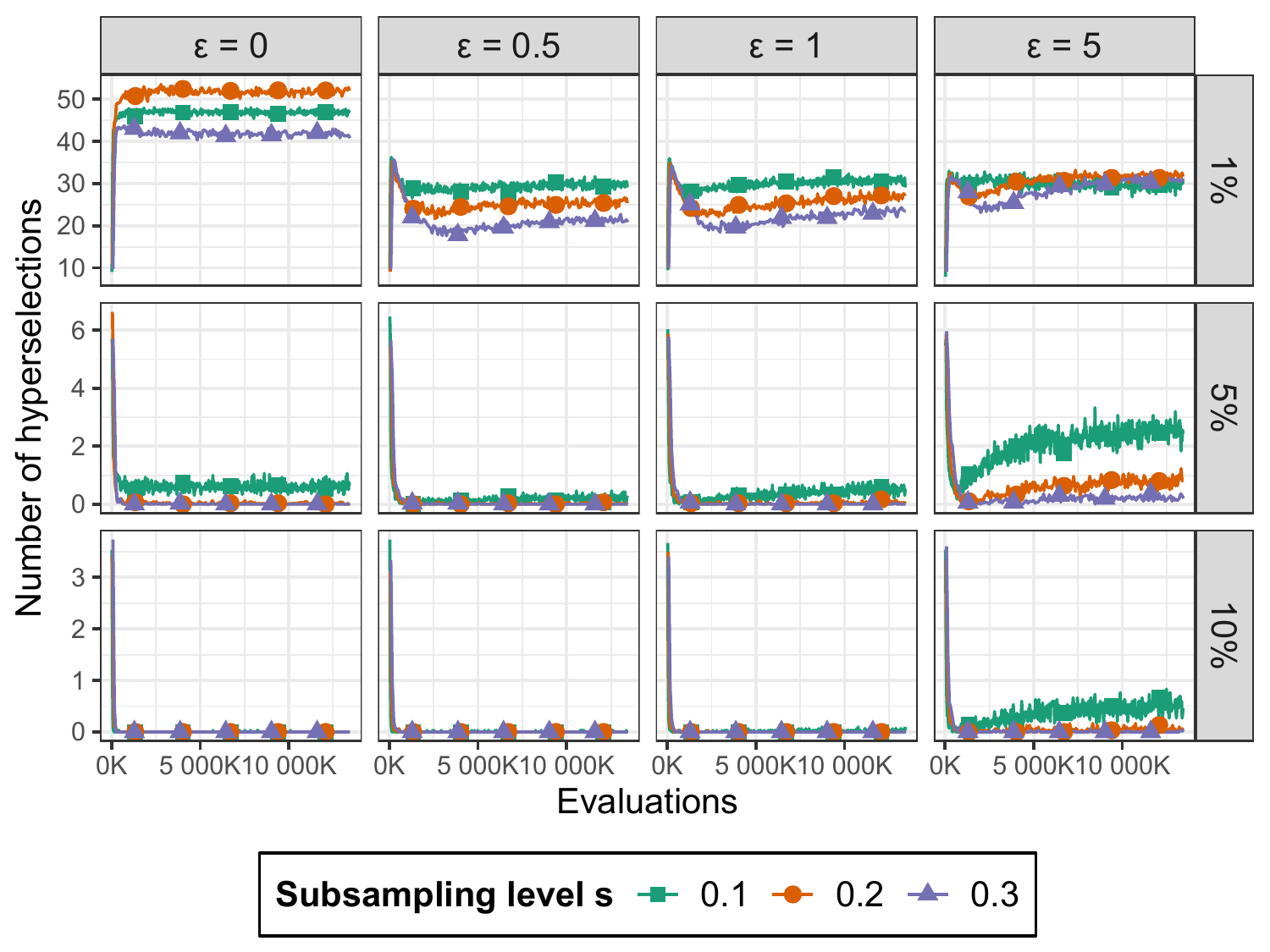}
    \caption{The number of hyperselections at different levels for different parameter settings of down-sampled $\epsilon$-lexicase selection for the ENC problem, shown over the number of evaluations. The results are averaged over 30 runs.}
    \label{fig:hyper_sub_param_enc}
\end{figure}

Besides the influence on performance, the parameter settings also have an effect on the properties of the population. Figure \ref{fig:sub_param_div_enc} shows the evolution of diversity for different parameter settings of down-sampled $\epsilon$-lexicase selection. We can see that a smaller subsampling level reduces diversity. 
If we look at Fig.~ \ref{fig:sub_param_div_enc} and Fig.~ \ref{fig:hyper_sub_param_enc}, we can see that the hyperselection rates correlate with diversity.  
This is not surprising, as hyperselection indicates how often the same individuals are selected. If the same individuals are selected more frequently this consequently lowers diversity.  
For example, for an $\epsilon = 0$ and an $\epsilon = 5$, the number of hyperselections at the 5\% level is largest for a subsampling level $s=0.1$ (see Fig.~\ref{fig:hyper_sub_param_enc}). This correlates with a low diversity for a subsampling level $s=0.1$ (see Fig.~\ref{fig:sub_param_div_enc}) as the same individuals are selected more often compared to a greater subsampling level. 
We assume that the reason for the larger hyperselection rates with lower subsampling level is due to the fact that less training cases are available in each generation.

\begin{figure}[h]
    \centering
    \includegraphics[scale=0.555]{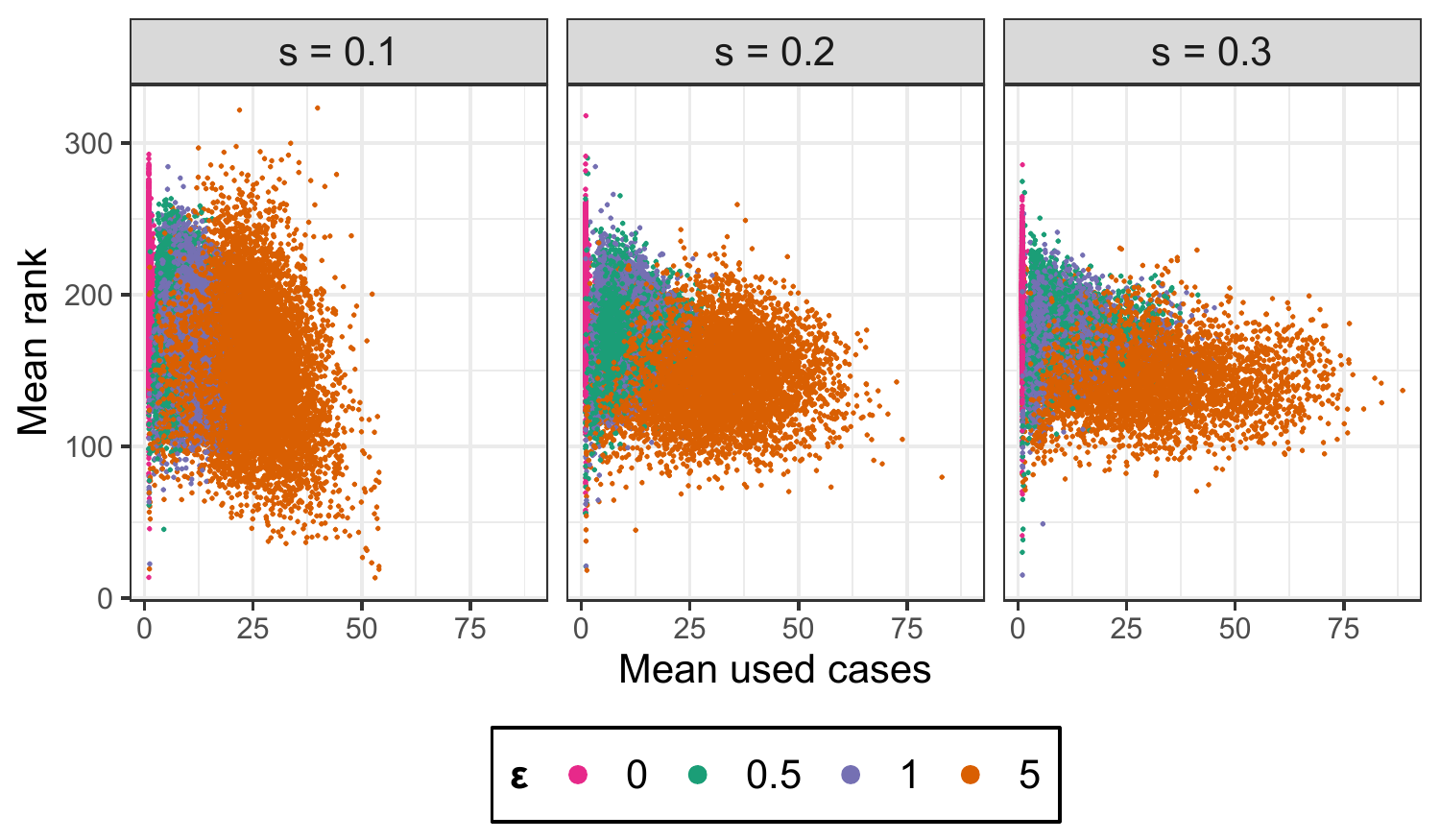}
    \caption{Mean rank of the selected individuals and mean number of used cases for selection per generation shown for different parameter settings of down-sampled $\epsilon$-lexicase selection for the ENC problem.}
    \label{fig:rank_cases_enc_sub}
\end{figure}

\begin{figure*}[h]
    \centering
    \includegraphics[scale=0.58]{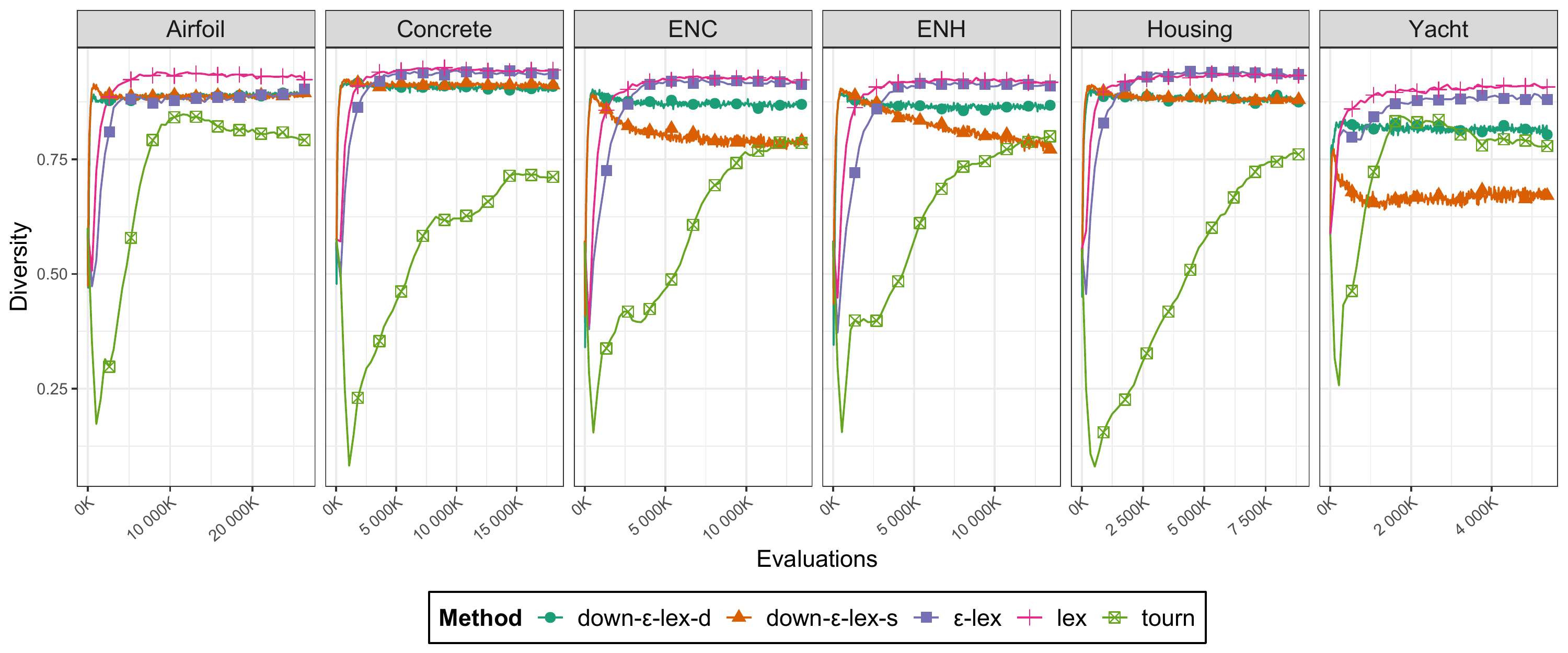}
    \caption{Behavioral diversity over the number of evaluations for all problems. The results are averaged over 30 runs. \vspace{1.2cm}}
    \label{fig:behavior_diversity}
\end{figure*}

\begin{figure*}[h]
    \centering
    \includegraphics[scale=0.58]{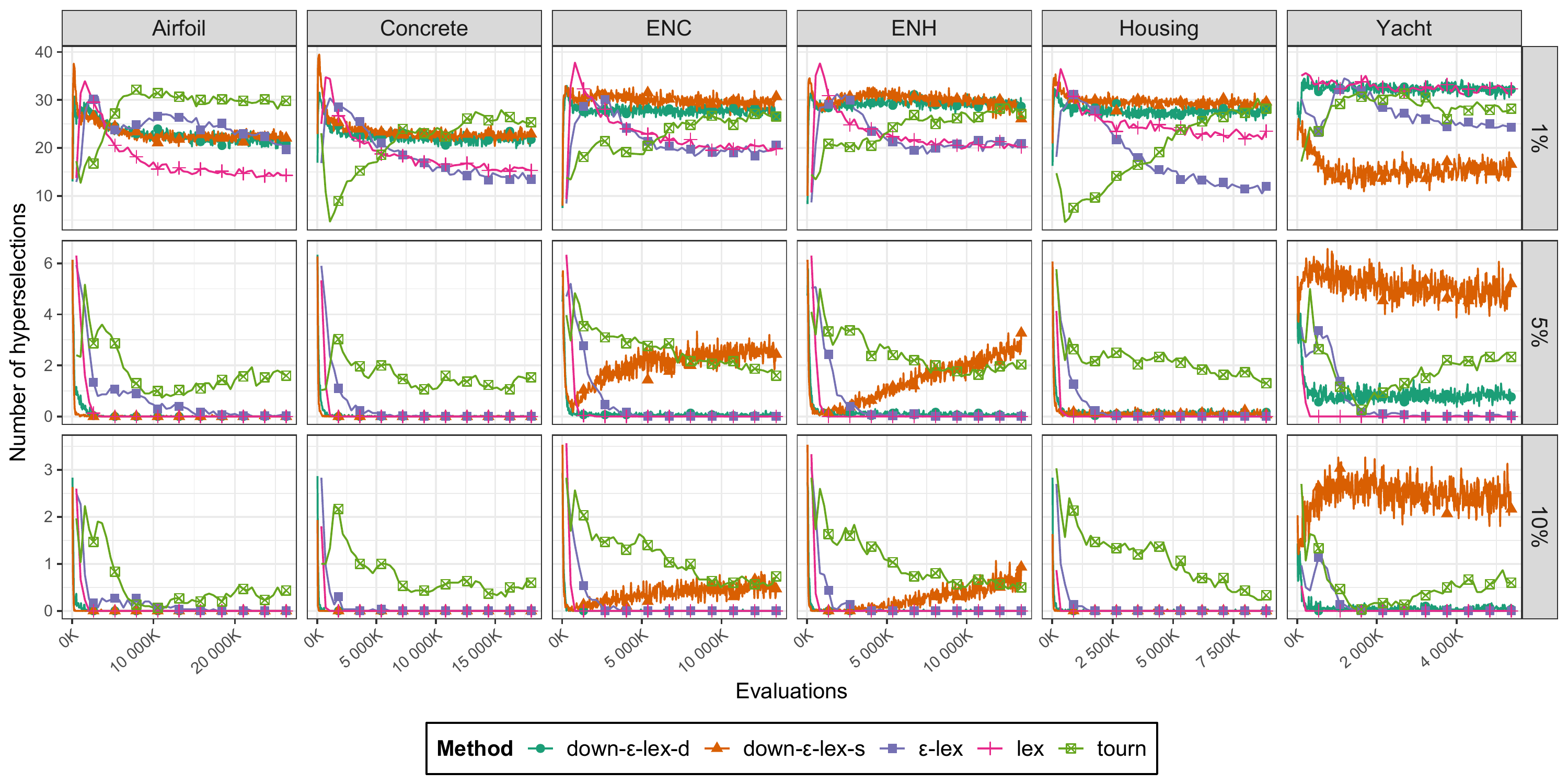}
    \caption{Number of hyperselection events over the number of evaluations for all problems. The results are averaged over 30 runs. \vspace{1.2cm}}
    \label{fig:hyperselection}
\end{figure*}

\begin{figure}[b]
    \centering
    \includegraphics[scale=0.555]{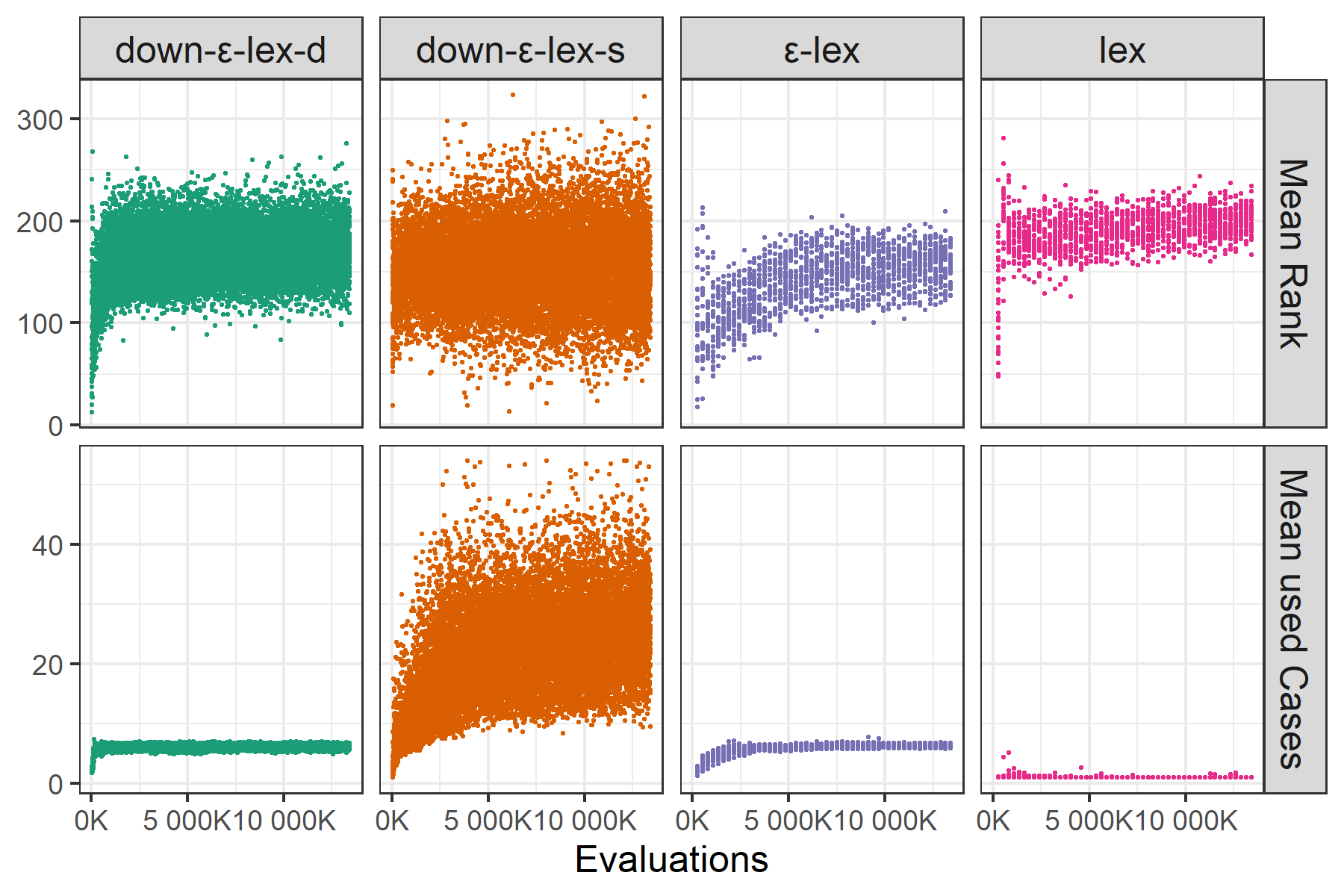}
    \caption{Mean number of training cases used for selection and mean rank of the selected individuals in each generation under each selection method, shown over the number of evaluations for the ENC problem.}
    \label{fig:mean_rank_cases_enc}
\end{figure}

\begin{figure*}
    \centering
    \includegraphics[scale=0.555]{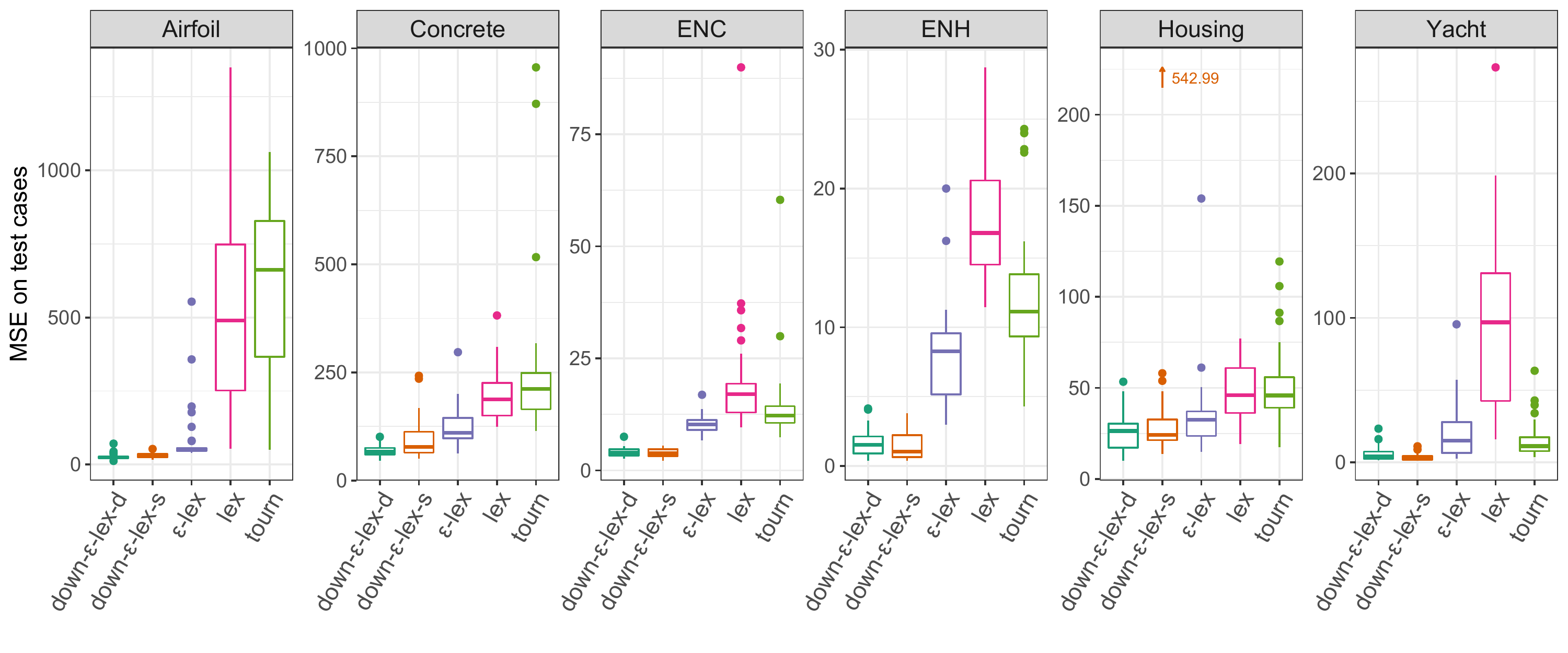}
    \caption{MSE on the test cases for all problems. On the Housing problem, the value of an extreme outlier is specified by an arrow.}
    \label{fig:all_results_boxplot}
\end{figure*}

\begin{table*}[h]
  \begin{center}
  \renewcommand{\arraystretch}{1.18}
    \caption{Median MSE on the test cases across all runs for all problems. The results are rounded to three decimal places. The best results are highlighted in bold font. Significant improvements with respect to the other selection methods are indicated by the method labels.}
    \label{tab:medianMSE}
    \begin{tabular}{lrrrrrr}
     \toprule 
      \textbf{Selection Method}\; & \textbf{Airfoil} & \textbf{Concrete}  & \textbf{ENC} &  \textbf{ENH} &  \textbf{Housing} & \textbf{Yacht} \\
      \midrule
      \rowcolor{lightgray}
      $^a$tourn & 661.809 & 211.915 & $^b$12.198 & $^b$11.138 & 45.876 & $^b$11.137 \\  
      $^b$lex &  489.910 & 187.328 & 17.012 & 16.795 & 45.969 & 96.964 \\
      
      \rowcolor{lightgray}
      $^c$$\epsilon$-lex &  $^{ab}$49.209 & $^{ab}$110.170 & $^{ab}$10.224 & $^{ab}$8.278 & $^{ab}$32.574 & $^b$15.052\\ 
          
      $^d$down-$\epsilon$-lex-s &  $^{abc}$29.693 & $^{abc}$77.413 & $^{abc}$\textbf{3.789} & $^{abc}$\textbf{1.049} & $^{ab}$\textbf{25.131} & $^{abce}$\textbf{2.688}\\ 

      \rowcolor{lightgray}
      $^e$down-$\epsilon$-lex-d  & $^{abcd}$\textbf{24.420} & $^{abcd}$\textbf{66.293} & $^{abc}$3.909 & $^{abc}$1.523 & $^{ab}$26.381 & $^{abc}$4.013 \\

      \bottomrule 
    \end{tabular}
  \end{center}
\end{table*}

Specialist selection has been found to influence the problem-solving performance of lexicase selection in the domain of program synthesis \cite{Helmuth.2020b}. A specialist is defined as an individual that has small errors on a small subset of training cases while having large errors on others \cite{Helmuth.2020b}. Therefore, we analyze the ability of down-sampled $\epsilon$-lexicase selection to select specialists, by observing the mean number of training cases used for selection and the mean rank of the selected individuals per generation. An individual with a poor rank that is selected using only a few cases is considered a specialist.
The case usage and the rank of the selected individuals is shown in Fig. \ref{fig:rank_cases_enc_sub} for different subsampling levels and different $\epsilon$ values. We can see that lower subsampling levels reduce the mean number of training cases that are used for selection. This can be explained by the fact that with a lower subsampling level, less training cases are available in each generation. In addition, we can observe that lower subsampling levels increase the variance in the average rank of the selected individuals. We assume that the reason for this is that smaller sample sizes lead to a greater variation between generations. In addition, Fig. \ref{fig:rank_cases_enc_sub} shows that larger $\epsilon$ values increase the case usage as more individuals pass each case in the selection process (see Alg. \ref{algo:down-ep-lex-math} line \ref{line:threshold}-\ref{line:pass_condition}). 

To sum up, we have found that a lower subsampling level leads to higher performance. We assume that the possibility to evaluate more individuals outweighs the lower population diversity observed with lower subsampling levels. 

\subsection{Comparison of Population Metrics}\label{sec:results_population_metrics}

To further analyze how down-sampled $\epsilon$-lexicase selection behaves, we compare it with traditional selection methods using the metrics introduced in Sect.~\ref{sec:metrics}. We choose parameter values for down-sampled $\epsilon$-lexicase selection based on the results shown in Sect.~ \ref{sec:param_setting}. We find that a smaller subsampling level performs better. Therefore, we use a subsampling level $s = 0.1$ to compare down-sampled $\epsilon$-lexicase selection with the traditional selection methods. In contrast, the optimal $\epsilon$ value is problem dependent \cite{LaCava.2016}. For the Concrete, ENH, and Housing problems, an $\epsilon=1$ performs best. For the Airfoil, ENC, and Yacht problems, the best results are achieved with an $\epsilon=5$. In addition, we show the results for down-sampled $\epsilon$-lexicase selection with a dynamically adapting $\epsilon$ that does not require the user to tune $\epsilon$. 

Figure \ref{fig:behavior_diversity} shows the evolution of behavioral diversity for tournament selection (denoted as tourn), lexicase selection (denoted as lex), $\epsilon$-lexicase selection (denoted as $\epsilon$-lex), down-sampled $\epsilon$-lexicase selection with a static $\epsilon$ (denoted as down-$\epsilon$-lex-s) and with a dynamic $\epsilon$ (denoted as down-$\epsilon$-lex-d) on all considered problems. Both down-sampled $\epsilon$-lexicase variants produce less diverse populations than $\epsilon$-lex on most problems. For the ENC and ENH problem, we can observe that diversity decreases over time using down-$\epsilon$-lex-s while it is constant using down-$\epsilon$-lex-d. For the Yacht problem, diversity is significantly lower using down-$\epsilon$-lex-s compared to down-$\epsilon$-lex-d. 

These results fit well to the results for the hyperselection rates shown in Fig.~\ref{fig:hyperselection}. We can observe that especially at a high hyperselection level the number of hyperselections negatively correlates with diversity as the same individuals are selected in a large percentage of the parent selection events in each generation. For example, for the Yacht problem, the number of hyperselections at the 5\% and 10\% level is largest for down-$\epsilon$-lex-s compared to the other selection methods which correlates with a low diversity. For the ENC and ENH problem, we can observe that the number of hyperselections at the 5\% and 10\% levels increase over time for down-$\epsilon$-lex-s which corresponds to the observed decrease in diversity.

Besides the influence on diversity, we analyze differences in the ability to select specialists between the selection methods. Figure \ref{fig:mean_rank_cases_enc} shows the mean number of used cases for selection and the mean rank of the selected individuals over the number of evaluations for the ENC problem. We only show the results for the lexicase-based methods, because tournament selection always uses all training cases anyway due to the use of an aggregated fitness value. We can see that $\epsilon$-lex and down-$\epsilon$-lex-d use around 6 cases on average for selection. In contrast, the case usage of down-$\epsilon$-lex-s increases over time. We assume that the reason for this is that the population converges, which means that more individuals pass the threshold in the selection process. This in turn increases the case usage \cite{LaCava.2016}.

\subsection{Performance Comparison}\label{sec:performance_comparison}

Finally, we compare the performance of down-sampled $\epsilon$-lexicase selection methods with the traditional selection methods. Therefore, we analyze the achieved MSE on the unseen test cases on all considered problems. To test whether the results significantly differ, we perform a pairwise Wilcoxon rank-sum test with a Holm correction following the approach of La Cava et al. \cite{LaCava.2016}. Significance is defined as $p < 0.05$. Table \ref{tab:medianMSE} shows the median MSE across all runs for all problems. Each method is assigned a label $a-e$ to indicate significant improvements with respect to the other selection methods. For the down-sampled $\epsilon$-lexicase variants, the results are shown for a subsampling level $s = 0.1$. 
To make the distribution of the results visible, Fig.~\ref{fig:all_results_boxplot} shows box-plots of the MSE on the unseen test cases.

We see that either down-$\epsilon$-lex-s or down-$\epsilon$-lex-d achieve the lowest median MSE on the test cases on all problems. For example, on the ENH problem, down-$\epsilon$-lex-s achieves a median MSE of $1.049$, while $\epsilon$-lex achieves only a median MSE of $8.278$.

Overall, down-$\epsilon$-lex-s and down-$\epsilon$-lex-d perform significantly better than tourn, lex, and $\epsilon$-lex on $5$ out of $6$ problems. On two problems, down-$\epsilon$-lex-d significantly outperforms down-$\epsilon$-lex-s. down-$\epsilon$-lex-s significantly outperforms down-$\epsilon$-lex-d on one problem. 

To sum up, we found that both down-sampled $\epsilon$-lexicase selection variants outperform traditional selection methods. We assume that the performance improvements are due the possibility to evaluate more individuals with the same evaluation budget as down-sampled $\epsilon$-lexicase selection searches for more generations than the other selection methods.

\section{Conclusions and Future Work}\label{sec:conclusion}

In this paper, we proposed down-sampled $\epsilon$-lexicase selection which combines subsampling and $\epsilon$-lexicase selection. To evaluate its performance, we compared down-sampled $\epsilon$-lexicase with traditional selection methods on common real-world symbolic regression benchmarks taken from the UCI machine learning repository. Furthermore, we analyzed the influence of down-sampled $\epsilon$-lexicase selection on the properties of the population over a GP run.

We found in our analysis that diversity is reduced by using down-sampled $\epsilon$-lexicase selection compared to standard $\epsilon$-lexicase selection. This correlates with high hyperselection rates using down-sampled $\epsilon$-lexicase selection. 
In addition, our results suggest that a smaller subsampling level performs better. Best results were achieved with a subsampling level $s = 0.1$. 

Further, we found that down-sampled $\epsilon$-lexicase selection outperforms the traditional selection methods on all studied symbolic regression problems. This applies to the variant with a static as well as to the variant with a dynamically adapting $\epsilon$. The improvement with respect to the median MSE on the unseen test cases is up to about 87\% in comparison to standard $\epsilon$-lexicase selection. Consequently, we recommend to use down-sampled $\epsilon$-lexicase selection as it often significantly improves the solution quality.  

In future work, we will study the influence of different population sizes and the number of generations on the performance of down-sampled $\epsilon$-lexicase selection.

\bibliographystyle{ACM-Reference-Format}
\bibliography{main}

\end{document}